\documentclass{article}
\usepackage[utf8]{inputenc}

\usepackage{authblk}

\usepackage{amsmath}
\usepackage{csquotes}
\usepackage{enumitem}

\begin{document}

\newcommand{\model}{m}
\newcommand{\inSet}{X}
\newcommand{\outSet}{Y}

\newcommand{\mail}{\text{M}}
\newcommand{\spf}{\text{S}}
\newcommand{\spfa}{\text{S}'}
\newcommand{\spam}{\texttt{spam}}
\newcommand{\nospam}{\texttt{no\_spam}}

\newtheorem{example*}{Example}

\providecommand{\keywords}[1]
{
  \small	
  \textbf{\textit{Keywords---}} #1
}

\title{Explaining Any ML Model? -- On Goals and Capabilities of XAI}
\author{Moritz Renftle}
\author{Holger Trittenbach}
\affil{Institute for Program Structures and Data Organization, Karlsruhe Institute of Technology (KIT), \textit{moritz@renftle.net, ht@neurocat.ai}}
\author{Michael Poznic}
\author{Reinhard Heil}
\affil{Institute for Technology Assessment and Systems Analysis, Karlsruhe Institute of Technology (KIT), \textit{michael.poznic@kit.edu, reinhard.heil@kit.edu}}
\setcounter{Maxaffil}{0}
\renewcommand\Affilfont{\itshape\small}
\date{June 2022}

\maketitle

\begin{abstract}
    An increasing ubiquity of machine learning (ML) motivates research on algorithms to \enquote{explain} ML models and their predictions -- so-called eXplainable Artificial Intelligence (XAI).
    Despite many survey papers and discussions, the goals and capabilities of XAI algorithms are far from being well understood. 
    We argue that this is because of a problematic reasoning scheme in XAI literature: XAI algorithms are said to complement ML models with desired properties, such as \emph{interpretability}, or \emph{explainability}. 
    These properties are in turn assumed to contribute to a goal, like \emph{trust} in an ML system.
    But most properties lack precise definitions and their relationship to such goals is far from obvious. 
    The result is a reasoning scheme that obfuscates research results and leaves an important question unanswered: What can one expect from XAI algorithms?
    In this article, we clarify the goals and capabilities of XAI algorithms from a concrete perspective: that of their users.
    \enquote{Explaining} ML models is only necessary if users have questions about them.
    We show that users can ask diverse questions, but that only one of them can be answered by current XAI algorithms.
    Answering this core question can be trivial, difficult or even impossible, depending on the ML application.
    Based on these insights, we outline which capabilities policymakers, researchers and society can reasonably expect from XAI algorithms.
\end{abstract}


\section{Introduction}
\label{sec:intro}
For long, achieving high predictive accuracy has been the lynchpin of machine learning (ML) research and a fundamental requirement for using ML in practice.
But an increasing pervasiveness of ML has led to concerns among policymakers, researchers and society about ethical and juridical implications of using ML without an \emph{understanding} of the behavior and limitations of the models generated~\cite{goodman_european_2017}.
This has fostered research, so-called eXplainable Artificial Intelligence (XAI), on algorithms that generate \enquote{explanations} of ML.
The explanations in turn are expected to improve the understanding of ML users and researchers. 

Over time, XAI algorithms have turned into an alleged panacea to address virtually any concern that may be raised in the context of ML.
There is a common scheme of reasoning regarding XAI algorithms.
First, one points out some important \emph{*ility}, i.e., a property that an ML model does not possess.
Typical properties of focus are interpretability, explainability and comprehensibility~\cite{szczepanski_methods_2021}.
The argument then goes that XAI algorithms supplement the *ility of interest through some form of explanation.
And this explanation in turn is expected to help achieving a postulated \emph{goal}, like building trust in a model~\cite{barredoarrieta_explainable_2020} or a justification for using the model in a specific context~\cite{adadi_peeking_2018}.

This scheme is problematic for various reasons.
One reason is an assumed relation between *ilities and goals that often is not well defined or characterized.
A second reason is that, in general, the *ilities are not sufficient and may not even be necessary to achieve all goals.
For instance, take the goal of trust and interpretability as the currently most cited *ility~\cite{lipton_mythos_2016,erasmus_what_2021}.
We argue that one can achieve trust in a technical system through reputation and past performance, and not necessarily by making the system interpretable.

But what seems to be an even more pressing problem is that there are diverse, contextual meanings of the terms used for these *ilities.
For example, there is no consensus on what interpretability means in the context of ML and which criteria an XAI method must fulfill to make an ML model interpretable; the same holds for most of the remaining *ilities.
This issue has been pointed out in the scholarly literature ever since~\cite{lipton_mythos_2016,krishnan_interpretability_2019,robbins_misdirected_2019,erasmus_what_2021}.

As a result of these problems, the common reasoning scheme seems insufficient to approach the goals and capabilities of XAI algorithms.
In this paper, we strive to clarify these two aspects.
Our approach is to take on the perspective of users of ML.
When users need a better understanding of ML models, this implies that they have some open questions about these models.
One can view XAI algorithms as methods that help users to answer these questions.
With this viewpoint, clarifying the goals 
of XAI algorithms means (i) to collect questions that users might have about ML models and (ii) to identify the subset of these questions that XAI algorithms help to answer.
Further, clarifying the capabilities of existing XAI algorithms means (iii) to examine what is hard about the questions identified in (ii), i.e., to identify the \emph{challenges} for these algorithms, and (iv) to assess how far the challenges are met by existing algorithms.

This article is structured accordingly.
In Section~\ref{sec:epistemic-questions}, we introduce several questions that users can ask about ML models.
In Section~\ref{sec:scope}, we find that only one of these questions is currently addressed by XAI algorithms.
Answering this question reveals two general challenges, the one of translation and the one of approximation, which we present in Section \ref{sec:challenges}.
In Section~\ref{sec:state-of-xai}, we review how far existing XAI algorithms meet the challenges.
Section \ref{sec:conclusions} concludes.

\section{Questions about ML Models}
\label{sec:epistemic-questions}
Users can ask many questions about ML models.
Here, we introduce seven questions that appear frequently in the literature (e.g.,~\cite{gunning_explainable_2017,hoffman_metrics_2018}).
We illustrate these questions with a use case of ML: the detection of spam emails through an ML model.
The questions will later serve to delineate the goals of XAI algorithms.
Additionally, while they are based on an example, we deem them quite general and applicable to other ML use cases as well.
\begin{example*}[Spam filter]
Assume a spam filter $\spf$ is an ML model that classifies emails into two disjoint groups, $\spam$ and $\nospam$.
Further, there is a user Alice who sees an email $\mail$ which $\spf$ classifies as $\spam$.
Alice is a curious mail recipient and has an interest in understanding the ML model.
She asks the following questions:
\begin{enumerate}[label=Q\arabic*:,ref=Q\arabic*]
    \item
        \label{q1}
        Why is $\mail$ classified as $\spam$ by $\spf$?
    \item
        \label{q2}
        How does $\spf$ distinguish $\spam$ from $\nospam$? 
    \item
        \label{q3}
        What distinguishes $\spam$ from $\nospam$?
    \item
        \label{q4}
        How does $\spf$ work?
    \item
        \label{q5}
        Does $\spf$ work like an alternative spam filter $\spfa$ she has used in the past?
    \item
        \label{q6}
        Alice thinks that $\mail$ is not spam. Why does $\spf$'s classification differ from her opinion? 
    \item
        \label{q7}
        What distinguishes spam emails from normal emails?
\end{enumerate}
\end{example*}

Let us now elaborate on these questions and clarify what is required to answer them.

\subsection*{\ref{q1}: Why is $\mail$ classified as $\spam$ by $\spf$?}
Answering this question requires to clarify what exactly $\spf$ is and on what basis it makes classifications.

Mathematically, $\spf$ is a function that maps from emails to either the label $\spam$ or $\nospam$.
Many such functions exist, but only few are an accurate spam filter.
To find one of these accurate functions, our example uses supervised ML.

Supervised ML is a general methodology for predicting a target variable $Y$ based on an observable variable $X$.
The goal is to find a function that maps values of the observable variable to values of the target variable, with low error.
To find such a function, supervised ML requires \emph{training data}.
Training data is a set of pairwise observations $(x_i, y_i)$ of $X$ and $Y$, where $i$ is an integer identifying the pair.
In our case, the $x_i$ represent emails, and the $y_i$ are either $\spam$ or $\nospam$.
$x_i$ can represent whole emails, e.g., as a sequence of zeros and ones in binary format.
But $x_i$ can also consist of only some attributes of an email, e.g., the domain of the sender or the number of words in the email body.
Supervised ML assumes that the training data is representative of the joint distribution of the two variables $X$ and $Y$.
There are different notions of representativeness, but the most frequent one is that (a) the observations in the training data have been sampled independently and identically distributed (i.i.d.) from $X$ and $Y$ and that (b) the number of observations is \enquote{large enough}.

Based on the training data, supervised ML selects a function to map from $X$ to $Y$ with low prediction error.
The function selection typically involves two steps: (i) choosing a set of candidate functions, and (ii) finding the optimal function within this set.
In step (i), one chooses a \emph{function type}, i.e., a set of functions whose equations have a similar form.
Examples of function types are linear functions, decision tree functions or neural network functions.
For the chosen function type, one further chooses so-called hyperparameters, e.g., the number of coefficients of a linear function.
The result of step (i) is a function with a number of free parameters, often called an \emph{ML model}.
In step (ii), one then uses an ML algorithm to optimize the remaining parameters of the model using the training data.
The algorithm returns a \emph{trained ML model}, i.e., the model together with the parameters that lead to the lowest prediction error for the observations in the training data.
If the training data is indeed representative of the joint distribution of $X$ and $Y$, then this model will also have a low prediction error for new observations of $X$ and $Y$.

In view of all this, we can now answer why $\spf$ makes particular classifications (\ref{q1}): the classifications of $\spf$ are the result of a specific supervised ML methodology.
This methodology involves many decisions and assumptions, e.g., regarding the collection of the training data or the selection of candidate ML models.
These decisions and assumptions determine both the overall prediction performance of $\spf$ and individual errors of $\spf$.

\subsection*{\ref{q2}: How does $\spf$ distinguish $\spam$ from $\nospam$?}
This question asks for the attributes of $\mail$ and the function by which $\spf$ distinguishes between the labels $\spam$ and $\nospam$.

An important distinction is between attributes that humans can perceive and interpret and attributes that refer only to the digital representation of a data item and exist only for the purpose of machine learning.
For brevity, we will call the former attributes \emph{interpreted attributes} and the latter \emph{technical attributes}.
In the case of an email, interpreted attributes include the sender domain, letters, words, phrases, or the word count.
A technical attribute would be, for example, the result of a principal component analysis in the form of numerical values~\cite{abdi_principal_2010}.
Technical attributes can lack human interpretation; e.g., one might not be able to tell what the first principal component means.
Another example for a technical attribute of emails is their representations as high-dimensional numerical vector embedding~\cite{le_distributed_2014}.

One can answer \ref{q2} in different ways.
A simple answer is to write down the whole ML model $\spf$, as a complex mathematical expression that involves technical attributes of emails.
But this answer has two problems.
First, $\spf$ might use many more numeric parameters and attributes of emails than a human mind can comprehend.
In this case, the mathematical definition of $\spf$ is a correct answer to \ref{q2}, but not one that satisfies users.
Second, even if the mathematical definition of an ML model is \enquote{simple}, it might still not satisfy users if it only refers to technical and not to interpreted attributes.
The reason is that users of $\spf$ typically have prior knowledge and expectations regarding the phenomenon \enquote{spam} they want to compare $\spf$ to.
This requires a definition of $\spf$ that ties to prior knowledge of users, i.e., that uses \emph{interpreted} attributes.
For example, if the function of $\spf$ relies only on the first principal component of an email vector embedding, it remains unclear whether this value is correlated with some interpreted attribute of emails.
Overall, a good answer to \ref{q2} for users would be a simple function that uses interpreted attributes of emails but still describes accurately how the spam filter $\spf$ classifies emails.
Note that such a function need not exist, or that there may be several functions that describe the original ML model with similar accuracy.

\subsection*{\ref{q3}: What distinguishes $\spam$ from $\nospam$?}
So far, we have focused on $\spf$ and classification.
\ref{q3} now changes the perspective.
The answer to what discerns $\spam$ from $\nospam$ is independent of $\spf$ and also independent of any other ML model; it depends only on the training data of~$\spf$.

A clarification of \ref{q3} is: \enquote{Which functions exist that distinguish the emails labeled $\spam$ from those labeled $\nospam$ within the training data?}.
As discussed before, users require functions that are simple and use interpreted attributes.
Hence, we only consider such functions in the following discussion.

The complete answer to \ref{q3} is to specify \emph{all} simple functions that use interpreted attributes and accurately distinguish $\spam$ from $\nospam$ within the training data.
Typically, the number of such functions may be very large or even infinite, depending on the type and number of attributes in the training data.
Thus, examining all these functions is infeasible for users.

So a less accurate but more useful answer to \ref{q3} is a small, diverse set of functions that are simple, use interpreted attributes and accurately distinguish $\spam$ from $\nospam$ within the training data.
There are two ways to generate such a set.
First, one can create new training data that only uses interpreted attributes of emails.
One then runs an ML algorithm that optimizes a simple function for distinguishing $\spam$ from $\nospam$ based on these attributes.
One repeats the process with other ML algorithms until one has a diverse set of functions.
Second, one can also start with a complex ML model, like $\spf$, and then derive from it different simple functions that use interpreted attributes.
This is equivalent to answering \ref{q2}.
One can further diversify the results by deriving the functions from different complex ML models.

\subsection*{\ref{q4}: How does $\spf$ work?}
This question asks for the mechanics of $\spf$.
A simple answer is that $\spf$ is a function that maps from emails to labels.
But one can also interpret this question as asking for the mechanism of how to derive $\spf$ from data.
With this interpretation, the answer to \ref{q4} includes the answer to \ref{q1}, i.e., the selection of a predictive function based on training data.
A full answer to \ref{q4} further requires a description of the complete supervised ML procedure and the algorithms that have been used to optimize $\spf$.
This description can be pseudo-code of the algorithms or a mathematical description.
In contrast, it is irrelevant for \ref{q4} how exactly the algorithms are implemented, i.e. in which programming language, with which libraries, etc., since this has no effect on their functionality.

\subsection*{\ref{q5}: Does $\spf$ work like an alternative spam filter $\spfa$?}
This question aims at comparing two trained ML models that use the same set of input attributes to compute their outputs.

Suppose that $\spf$ and $\spfa$ are white boxes, i.e., one knows their function types and parameters.
Then $\spf$ and $\spfa$ work equivalently if they use the same function type and have identical parameters.
Note that the inverse is not true, i.e., two models can use different function types and parameters and still represent the same mathematical function.
In this case, showing the equivalence of the two models is much more difficult.

However, if either $\spf$ or $\spfa$ is a black box, one does not know any explicit representation of its function.
In this case, one can only make equivalence statements with respect to a set of observations.
For example, one can verify that $\spf$ and $\spfa$ yield the same classification for a set of emails.
This does not show, however, that the functions of $\spf$ and $\spfa$ are identical -- their results could differ on other observations.

Summing up, answering question \ref{q5} is simple if the answer is \emph{no}: one only needs one email where $\spf$ and $\spfa$ yield different classifications.
In contrast, proving that the answer to \ref{q5} is \emph{yes} is hard or almost impossible -- except in the trivial case that both models are white boxes with the same function type and parameters.

\subsection*{\ref{q6}: Why does the classification of $\spf$ differ from Alice's opinion?}
This question aims at comparing Alice's personal reasoning with the classification function $\spf$.
Answering the question requires to investigate whether $\spf$ relies on the same attributes of emails as Alice does.
If this is not the case, one already has an answer: $\spf$ relies on different attributes.
If $\spf$ and Alice rely on the same attributes, one must further compare the function $\spf$ to the one of Alice, i.e., whether the attributes are used in the same way.
We have already discussed this in the answer to \ref{q5}.

\subsection*{\ref{q7}: What distinguishes spam emails from normal emails?}
This question asks for a universal rule.
If a commonly accepted definition of spam exists, then the answer is simple: if an email meets the definition of a spam email, then it is spam. 
If the definition does not exist, one can take a set of examples of spam and normal emails and develop an ML model with it.
The question then is whether the trained ML model is helpful in providing an answer to \ref{q7}.
This depends on the degree of consensus among email recipients about what does and does not qualify as spam email.
If all recipients agree with each other, then any trained ML model with a high accuracy is a possible answer to \ref{q7}; and any such ML model could help to establish a commonly accepted definition of spam.
However, if the recipients disagree on which email is spam or not, there can be no universal definition of spam, and \ref{q7} is not answerable.

\section{Questions Addressed by XAI Algorithms}
\label{sec:scope}
In the following, we will show that existing XAI algorithms do not give answers to any of the questions introduced in the previous section but one: \ref{q2}.
Each of the following paragraphs will make an observation about XAI algorithms that rules out some of the questions, until only question~\ref{q2} remains.
We will then examine \ref{q2} closer and show the centrality of this question to present XAI research.

\paragraph{Real-world phenomena}
There is consensus in the literature that XAI algorithms primarily explain ML models and not real-world phenomena~\cite{gunning_explainable_2017,samek_explainable_2017,adadi_peeking_2018,barredoarrieta_explainable_2020}. 
Question \ref{q7} asks for a characterization of a real-world phenomenon: spam.
Although one can use ML models to explore possible answers to questions like \ref{q7}~\cite{sullivan_understanding_2019}, the decision of what is the correct answer is independent of the models.
Hence, \ref{q7} is not addressed by XAI algorithms.

\paragraph{Creation process of ML models}
XAI algorithms so far tend to generate \enquote{explanations} of ML models that are independent of how the models were created.
To illustrate, consider a popular family of algorithms, so-called \emph{feature attribution} methods, e.g.,~\cite{ribeiro_why_2016,sundararajan_axiomatic_2017}.
Feature attribution methods estimate, for a given input to an ML model, how important the individual features of the input are for the corresponding output of the model.
One can compute such feature attributions without knowing how the model was created.
In contrast, knowledge of the creation process of the ML model $\spf$ is necessary to answer how $\spf$ distinguishes $\spam$ from $\nospam$ (\ref{q1}), and how $\spf$ works (\ref{q4}).
We conclude that \ref{q1} and \ref{q4} are, at present, not answered by XAI algorithms.

\paragraph{Complex ML models}
XAI research mainly focuses on ML models that are very complex, e.g., because they have millions of numeric parameters.
In this setting, it is highly improbable that two ML models created in different ways end up with the same parameters and hence implement the same function.
Thus, for the ML models considered by XAI research, questions \ref{q5} and \ref{q6} have trivial answers: Complex ML models hardly ever work in the same way; and the classification function of a user is expected to differ from the one of a given complex ML model.

\paragraph{\ref{q3} reduces to \ref{q2}}
Recall that we have identified two ways to answer what distinguishes $\spam$ from $\nospam$ (\ref{q3}).
The first way applies traditional supervised ML methodology, namely feature engineering and the training of simple types of ML models.
This methodology has been established long before XAI emerged as a research area.
The second way of answering \ref{q3} is to develop one or more complex ML models, like $\spf$, and then derive from them diverse simple functions that use interpreted attributes.
This is equivalent to collecting different answers to question \ref{q2}.
Thus, if XAI algorithms address \ref{q3}, then only partially, and only as a side effect of addressing question~\ref{q2}.

\medskip
After these observations, only question~\ref{q2} is left for discussion: how does $\spf$ distinguish $\spam$ from $\nospam$?
For generality, we now decouple \ref{q2} from our example ML model $\spf$.
The generalized question is:
\begin{enumerate}[label=$\text{Q}^{*}$:, ref=$\text{Q}^{*}$]
    \item
        \label{q*}
        How can one represent an ML model as a simple function that uses interpreted attributes?
\end{enumerate}
From our perspective, this is the core question addressed by XAI algorithms.
Almost all existing XAI algorithms contribute in some way to approximating complex ML models with simpler functions, e.g., \cite{guidotti_survey_2019}, or with functions that use interpreted instead of technical attributes, e.g., \cite{bau_network_2017,kim_interpretability_2018,linardatos_explainable_2021}.
Considering the ambiguity of terms that the literature usually relies on for describing the capabilities of XAI algorithms -- i.e., \emph{interpretability}, \emph{explainability}, and so on -- it seems surprising that one can accurately describe these capabilities with one simple question.
We have the impression that literature and debates often make normative claims on XAI algorithms that go far beyond the question \ref{q*}.
For example, it is far from clear how \ref{q*} relates to terms like \enquote{trust}, or \enquote{fairness}, which carry normative connotations.
While these terms may relate to XAI algorithms in one or another way, using them without an explicit and shared interpretation obfuscates the results of XAI research.
It spurs misconceptions on the capabilities of XAI algorithms in politics and society, and ultimately harms the credibility of ML research.
So we advocate a more realistic approach to describe goals and capabilities of XAI algorithms: by focusing on the questions they can currently address.

\section{Challenges for XAI Algorithms}
\label{sec:challenges}
Given the core question~\ref{q*} addressed by XAI algorithms, we now examine what is hard about answering it.
We identify two main challenges.

\subsection{Approximation}
The first challenge is to approximate complex ML models with simpler functions, so-called \emph{surrogate models}%
\footnote{A technical remark: Our broad definition of \enquote{surrogate model} includes approaches such as feature attribution or counterfactuals, because one can view them as local approximations of an ML model.
The usage of this term may differ in other literature on XAI.}.
An elementary way to obtain a surrogate model is to observe a large set of inputs and corresponding outputs of the complex ML model and then train a simple ML model on these observations.
That is, the surrogate model is optimized to predict the outputs of the complex model, and not to predict the \enquote{true} outputs.
The accuracy of the surrogate model with respect to the complex model, i.e., how well the surrogate predicts the complex model, is sometimes called \emph{fidelity}~\cite{guidotti_survey_2019}.

Approximating ML models is both a technical and a conceptual challenge.
For instance, there is a tradeoff between the fidelity and the complexity of surrogate models: 
Simpler surrogate models will, in general, achieve lower fidelity than complex ones.
Thus, which surrogate model is optimal for users depends on how the users balance fidelity and complexity.

Further, there are many definitions of the complexity of an ML model.
An elementary one is the number of computational steps that one must perform to obtain the output of the model for a given input.
But there are also definitions that are specific to types of ML models.
For instance, one often quantifies the complexity of decision trees, a specific type of ML models, with metrics that are specific to tree structures, e.g., the length of the longest path from the root to a leaf node.
The lack of a universal definition of model complexity can make it difficult to choose between different types of surrogate models.

\subsection{Translation}
The second challenge in answering \ref{q*} is to translate the technical attributes that ML models use to discriminate between data items to interpreted attributes.
With technical attributes we mean both the inputs of an ML model and the intermediate results that the model computes for its prediction.
Interpreted attributes are all attributes of data items that humans can perceive, interpret, and communicate to each other using common terminology.

The difficulty of the translation challenge depends on the design and application of the ML model.
For some models and applications, it is clear how interpreted attributes relate to the technical attributes used by the model.
For example, suppose that the spam filter $\spf$ uses a digital representation of emails that consists of a single bit, i.e., a $0$ or $1$.
Suppose further that this bit is set if and only if an email contains the word \enquote{money}.
In this case, the technical attribute has a trivial mapping to an interpreted attribute.
In other cases, one might not know this mapping, although it exists.
For example, suppose that $\spf$ relies only on the first principal component of an email vector embedding, and that the value of this principal component correlates strongly with the presence of the word \enquote{money}.
One does not know this correlation until investigating it.
Finally, for some models and applications there does not exist any terminology to accurately describe the inputs or the intermediate results of the models.
In the previous example, this is if the principal component used by the ML model does not correlate with any already known term used to describe emails.
In summary, the translation challenge can be trivial, hard, or even unsolvable, depending on the ML application.

\section{The State of XAI Algorithms}
\label{sec:state-of-xai}
We now assess how far existing XAI algorithms address the approximation and translation challenges raised by question~\ref{q*}.
This section is not a comprehensive review, but rather a summary of present XAI algorithms with some examples.

There is a plethora of XAI algorithms that address the approximation challenge, for different types of complex ML models and surrogate models, and for different definitions of complexity and fidelity.
To give some examples, some algorithms can approximate complex ML models, like neural networks and random forests, with simpler models, like decision trees~\cite{craven_extracting_1996,bastani_interpreting_2019} or rule lists~\cite{benard_interpretable_2021}.
There are also algorithms to approximate complex ML models locally, i.e., in the vicinity of a given input.
Popular examples for this are LIME~\cite{ribeiro_why_2016}, SHAP~\cite{lundberg_unified_2017} and integrated gradients~\cite{sundararajan_axiomatic_2017}.
See~\cite{guidotti_survey_2019} for a comprehensive overview on approximation methods.

Compared to the approximation challenge, the translation challenge has received less attention.
Some work addresses ML models for computer vision, i.e., convolutional neural networks (CNNs) that make predictions based on images.
There are XAI algorithms to discover correlations of neurons or layers in CNNs with interpreted attributes of images, say, visible objects, shapes or colors~\cite{zhou_comparing_2019,kim_interpretability_2018,cammarata_curve_2020,goh_multimodal_2021}.
The discovered correlations enable users to map technical attributes used by CNNs, i.e., neuron or layer activations, to interpreted ones.
The main limitation of the methods is that they require a comprehensive database of annotated concepts on images.
To address this issue, \cite{ghorbani_automatic_2019} extract concepts in an automated way.
Experiments in this work indicate that many technical attributes of images used by neural networks can be mapped to interpreted ones.
Similar to these examples from computer vision, there also is work on translating ML models for natural language processing~\cite{poerner_interpretable_2018} and speech recognition~\cite{krug_neuron_2018}.

Answering question~\ref{q*} often poses a combination of the approximation and translation challenges.
Think of a typical computer vision model, an image classifier.
The classifier uses technical attributes of images, typically numeric pixel values for each color channel of the image.
Now suppose that one approximates the classifier with a surrogate model.
The surrogate model will either have low fidelity, or it will be almost as complex as the classifier.
The reason is that image classification is an inherently complex task: 
Functions that map from pixel-wise color values to the target classes are in most cases either complex or not accurate.
However, an answer to \ref{q*} can ignore some of this inherent complexity.
This is because users reason about images with interpreted attributes, e.g., shapes or objects.
Answering question~\ref{q*} does not require to approximate any parts of an image classifier that serve only to reconstruct interpreted attributes from pixel values.
On the contrary: a better answer to \ref{q*} is one that (i) translates technical attributes in the classifier to interpreted ones and (ii) defines an approximation of the classifier based on the translated attributes.

Such joint translation and approximation of ML models is an open challenge for XAI algorithms.
It occurs in all applications of ML models where the inputs have a lower level of abstraction than the terminology that users rely on.
Besides computer vision, this also holds for speech recognition and natural language processing.
Another example is the application of ML models for predicting physical processes; e.g., ML models that derive their predictions from the states of individual atoms, whereas their users observe aggregated changes on the macro level of the material under study.
One way to address the joint approximation and translation challenge is \emph{neurosymbolic AI}, i.e., \enquote{to develop neural network models with a symbolic interpretation}~\cite{garcez_neurosymbolic_2020}.
A pioneering example is the Neural Prototype Tree~\cite{nauta_neural_2021}, i.e., a neural network that learns a few \enquote{prototypical} combinations of interpreted attributes and then classifies inputs based on their resemblance to these prototypes.

\section{Conclusions}
\label{sec:conclusions}
A problematic reasoning scheme in the literature currently obfuscates both the general goals and capabilities of existing XAI algorithms.
In this paper, we have explored another way to characterize XAI algorithms, from the perspective of their users.
We have found that current XAI algorithms mainly address one particular question that users have in the context of ML: how can one represent a complex ML model as a simple function of interpreted attributes?

The succinctness of this question contrasts with the ambiguity of terms used in the literature to characterize XAI algorithms, such as \emph{interpretability} and \emph{explainability}.
Other terms in the literature have a normative connotation, e.g., \emph{trust}, whereas the identified question is purely technical.
The contrast between prevalent terminology and actual goals of XAI algorithms may spur excessive expectations of the algorithms on the part of policymakers and society.
To avoid this, we propose to focus on questions that the algorithms can actually help with.
Our characterization of XAI algorithms further reveals two key challenges for XAI research: the approximation and translation of ML models.
For approximation, the literature already offers many approaches; slightly fewer approaches exist for translation.
We think that holistic methods that address both challenges will be key in future research.

\bibliographystyle{abbrv}
\bibliography{xai_capabilities.bib}

\end{document}